\documentclass{article}
\usepackage{ijcai18}

\usepackage{times}
\usepackage{xcolor}
\usepackage{soul}
\usepackage[utf8]{inputenc}
\usepackage[small]{caption}
\usepackage{epsfig}
\usepackage{amsmath}
\usepackage{amssymb}
\usepackage{hyperref}
\title{MIXGAN: Learning Concepts from Different Domains for Mixture Generation}
\author{Guang-Yuan Hao$^1$, Hong-Xing Yu$^{1,4}$, Wei-Shi Zheng$^{1,2,3}$
\thanks{Corresponding Author}
\\
$^1$ School of Data and Computer Science, Sun Yat-sen University, China \\
$^2$ Key Laboratory of Machine Intelligence and Advanced Computing, Ministry of Education, China \\
$^3$ Collaborative Innovation Center of High Performance Computing, NUDT, China 
\\ 
$^4$ Guangdong Key Laboratory of Big Data Analysis and Processing, Guangzhou, China\\
guangyuanhao@outlook.com,
xKoven@gmail.com,
wszheng@ieee.org}

\begin{document}
% \newgeometry{top=6cm,bottom=1cm}

\onecolumn{

\noindent \textbf{MIXGAN: Learning Concepts from Different Domains for Mixture Generation}

\vspace{2cm}

\noindent {\LARGE{Guang-Yuan Hao, Hong-Xing Yu, Wei-Shi Zheng}}

\Large
\vspace{2cm}

 \noindent Code is available at the project page: \url{https://github.com/GuangyuanHao/MIXGAN}

\vspace{1cm}

\noindent For reference of this work, please cite:

\vspace{1cm}
\noindent MIXGAN: Learning Concepts from Different Domains for Mixture Generation
\\ Guang-Yuan Hao, Hong-Xing Yu, Wei-Shi Zheng, IJCAI, 2018

\vspace{1cm}

\noindent Bib:\\
\noindent
@article\{hao2018mixgan,\\
\ \ \   title=\{MIXGAN: Learning Concepts from Different Domains for Mixture Generation\},\\
\ \ \  author=\{Guang-Yuan Hao, Hong-Xing Yu, Wei-Shi Zheng\},\\
\ \ \  journal=\{ International Joint Conference on Artificial Intelligence\},\\
\ \ \  year=\{2018\}\\
\}
}
% \clearpage
% \restoregeometry
\newpage

\maketitle
\begin{abstract}
In this work, we present an interesting attempt on mixture generation: absorbing different image concepts (e.g., content and style) from different domains and thus generating a new domain with learned concepts. In particular, we propose a mixture generative adversarial network (MIXGAN). MIXGAN learns concepts of content and style from two domains respectively, and thus can join them for mixture generation in a new domain, i.e., generating images with content from one domain and style from another. MIXGAN overcomes the limitation of current GAN-based models which either generate new images in the same domain as they observed in training stage, or require off-the-shelf content templates for transferring or translation. Extensive experimental results demonstrate the effectiveness of MIXGAN as compared to related state-of-the-art GAN-based models.

\end{abstract}

\section{Introduction}

\begin{figure}[]
\begin{center}
\includegraphics[width=0.85\linewidth]{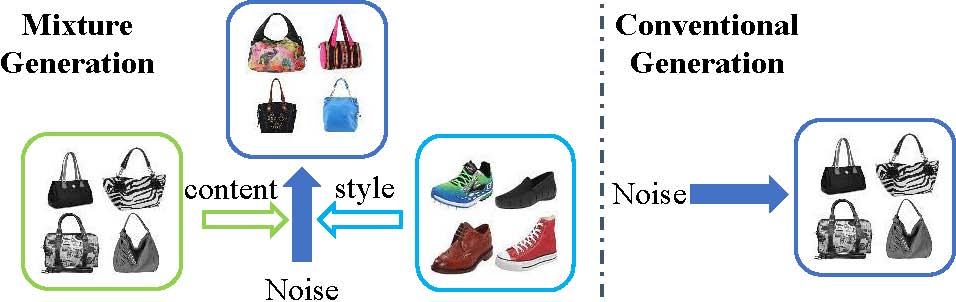}
\end{center}
   \caption{Mixture generation and conventional generation. Mixture generation requires absorbing concepts from different domains. In particular, here we learn content concept from the bags and style concept from the shoes for mixture generation. In contrast, conventional generation does not go beyond the available training domain.}
\label{fig:intro}
\end{figure}

 When you are looking at a red T-shirt in eBay, you could easily imagine how you would look like when you wear it: you know well the shape of your body, you have an image of this red T-shirt in your mind, and thus you can wear it in your imaginary world. However, can a learning machine do a job like this? This means that the machine should have the ability to learn from different domains (people and T-shirts, in the running example) and extract some specific concepts from them, respectively (people’s body shapes and T-shirts’ color style). Then, it is expected to join the specific kinds of concepts and thereby generate a new domain (imagination on wearing the T-shirt).  In realistic applications, this generation strategy might be more desirable in some scenarios other than the conventional generation strategy where only one training domain exists and the generated domain is expected to be the same as the existing one. We show another example in Figure \ref{fig:intro}. Here we aim to generate images in a new domain (i.e. colorful bags), given the content of one domain (the shape of the bags) and style of another domain (the color style of the shoes). By this way, it helps bring new ideas and provides visualizations for the bag designers who only have some raw ideas about designing bags of different color styles from the existing ones.

% image for introduction and showing our model's differences with traditional models 
% \label{fig:intro}
%,height=0.36\linewidt

We recognize such problems as the \emph{mixture generation} problems, where we need to \emph{jointly} absorb different kinds of concepts for generating a new domain beyond the available ones. \footnote{Here we interpret the mixture of concepts in a simplified way. For better understanding we refer the readers to the materials about \emph{conceptual blending} \cite{new1,new2}.}
Unfortunately, as illustrated in the right part in Figure \ref{fig:intro}, existing GAN-based {generative} methods are restricted to generate new samples similar to the ones from training domains \cite{gan,aae,acgan,began}. On the other hand, although the style transfer \cite{c9,c19,c10} and image-to-image translation \cite{c15,c39,c20} models can translate an existing image to another style, they are also restricted in that they require off-the-shelf content templates, and thus cannot deal with the problems requiring going beyond the available content templates (e.g., designing new shapes as well as styles of bags). Therefore, the mixture generation problem still remains an open issue.

To explicitly learn different types of concepts from different domains, in this work we focus on learning content (e.g., shape of bags) from one domain, and style (e.g., color style of shoes) from another. The main idea is that, in the generation process, the style concept ``joins'' the content concept, so that a generator can not only keep the content concept in mind, but also absorb the style concept to generate images of a new domain. To this end, we develop a framework in which the content concept is represented as hierarchical features, while the style concept is learned and embedded in a hierarchical decoder. During generation, the hierarchical features (with content concept) progressively passes through the decoder (with style concept), and meanwhile the decoder ``releases'' the style concept. Thereby, the style concept joins the content concept and finally the decoder produces an image based on them. As an analogy which might not be very precise but helps to understand, one can imagine barbecuing. We have pork and fish at hand (content), put them on the grill (decoder), spread oil on them at an early stage and sprinkle spice right before enjoying them (progressively absorbing the style concept). Also note that in the very beginning, we pay certain amount of money (noise for generation) to buy the pork and spice.

More specifically, we propose a \emph{mixture generator} model in our framework for mixture generation. The mixture generator consists of a content decoder and a mixture decoder. The content decoder learns from images of the content domain, and thus provides the mixture decoder with rich hierarchical features. These features, from mid-level to low-level, contain corresponding content concept and they are then progressively fed to the mixture decoder. The mixture decoder is designed to not only learn the style concept from the style domain, but also learn how to join both kinds of concepts for mixture generator.

With the mixture generator, we form a mixture generative adversarial network (MIXGAN). We evaluate our MIXGAN on several tasks. The experimental results show that our model can learn to generate images in a new domain, e.g.,  generating hand-written colorful digits provided that our model only observes black-and-white hand-written digits \cite{mnist} and colorful type-script ones \cite{svhn}.
The main contributions are as follows:

1.	We recognize the mixture generation problem that requires jointly absorbing concepts from different domains for generation, and propose to address it by joining style concept and content concept during generation.

2.	We propose an unsupervised framework as well as a novel model, i.e., mixture generator, to join the style and content concept for mixture generation. 
Specifically, the learned content concept is represented as hierarchical features and the style concept is embedded in a mixture decoder. During generation, the decoder ``releases'' the style concept which can thus be absorbed by the content concept.

3.	We show that our model can learn to generate images with content of one domain and style of another in several mixture generation tasks.

\section{Related Work}
 \paragraph{GAN-based generative models.} Generative Adversarial Nets (GAN) \cite{gan} is a popular framework for generation. It is like a two-player game, where a discriminator learns to distinguish real images from fake ones, while a generator tries its best to fool the discriminator. It is trained in an adversarial learning pattern, where the discriminator and generator iteratively improves themselves to beat the other one. Recently, lots of GAN-based models have largely promoted many generation fields, e.g., image generation \cite{gan,dcgan}, image editing \cite{c59}, and variational inference \cite{aae}, interpretable representation learning \cite{c38,cogan},etc.

Our work is closely related to GANs, as our framework adopts the adversarial learning pattern for training. On the other hand, our model is different from these models in that our model is designed for mixture generation, and therefore can generate a new domain. In contrast, these models are designed for conventional generation problems that do not require generating samples beyond training domains.

 \paragraph{Style Transfer and Image-to-Image Translation.}
Our work is also closely related to the style transfer models \cite{c9,c19,c10,domaintransfer} and image-to-image translation models \cite{c15,c39,c20}. Style transfer models aim to learn how to transfer the style of a source image to a target image \cite{c9}, while the image-to-image translation models learn a translation mapping from images of one domain to those of another domain \cite{pix2pix}. Typically the translation models learn the mapping from one style to another \cite{pix2pix}, e.g., translating a photo to a painting. Some image-to-image translation models adopt conditional generative adversarial network to learn a mapping from paired images, e.g. translating sketches to photos \cite{pix2pix,c8,c15,c19,c23,c28,c40,c50,c52,c55}. Recently, unpaired image translation also receives much research efforts \cite{cyclegan,discogan,dualgan,dtn,unit,unitgan,stargan}.

Our framework is related to them since our model and these models both need to learn style from a specific domain. However, the translation models translate images in one domain to another domain and the style transfer models learn how to transfer style in a pair of images, whereas our framework aims to learn concepts from different domains for mixture generation.

%, height = 0.4\linewidth
\begin{figure*}[t]
\begin{center}
\includegraphics[width=0.85\linewidth]{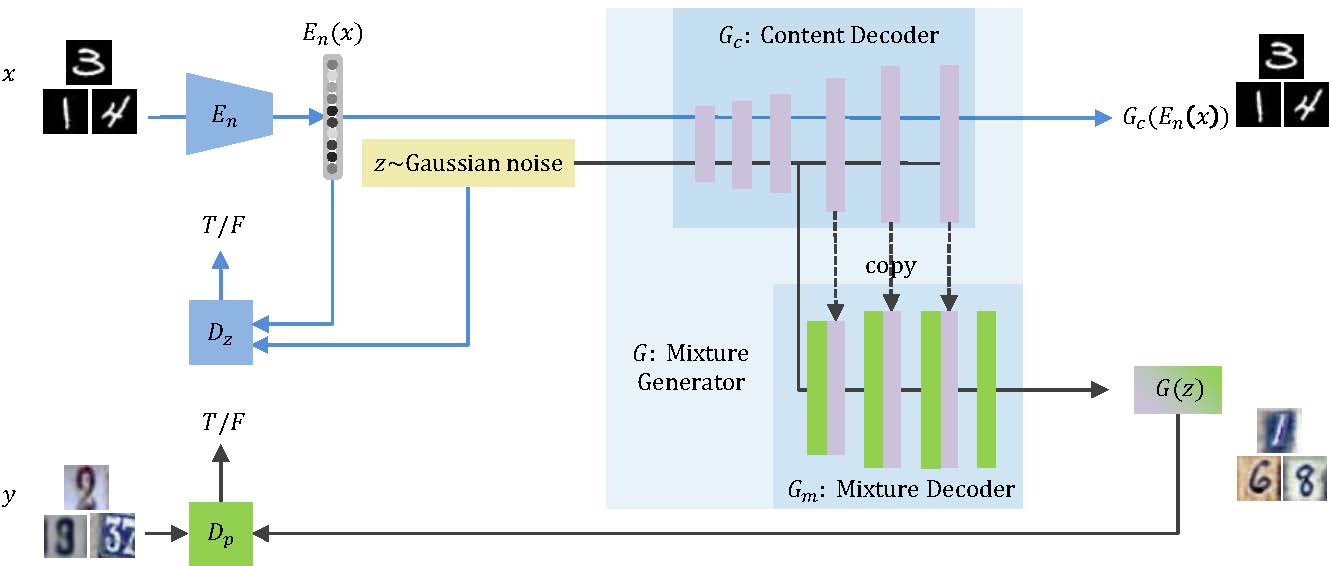}
\end{center}
   \caption{The structure of our mixture generative adversarial network (MIXGAN). The core of our framework is the mixture generator $G$, which consists of a content decoder $G_c$ and a mixture decoder $G_m$. The content decoder connects the content generation network (upper part in the figure) and the mixture generation network (mixture generator with a patch discriminator $D_p$) for joint learning, while the mixture decoder absorbs content concept of $x$ and style concept of $y$ for mixture generation. ($Best\ viewed\ in\ color.$)}

\label{fig:method}
\end{figure*}

\section{Framework}

In this section we present our framework. In the following, we will first give an overview of our framework, and then demonstrate how to learn content concept from a specific domain. Based on the content concept we develop the mixture generator which jointly learns the style concept and learns to join both kinds of concepts for mixture generation.

\subsection{Overview}
We propose an adversarial framework for jointly learning two kinds of concepts from two domains, respectively.
The general idea is to introduce a jointly learned mixture generator $G$ to absorb the two concepts. The mixture generator consists of a content decoder $G_c$ which learns the content concept, and a mixture decoder $G_m$ which jointly learns the style concept as well as how to join them for mixture generation.
The content decoder $G_c$ is learned within an adversarial autoencoder together with an encoder $E_n$ and a discriminator $D_z$. $G_c$ provides the mixture decoder $G_m$ with rich hierarchical information on the content concept, so as to connect the content generation and the style generation. $G_m$ is jointly learned with $G_c$ (for learning content concept) and a patch discriminator $D_p$ (for learning style concept). Thus, $G_m$ learns to join these two kinds of concepts. By this way, the two concepts are mutually learned and connected within the mixture generator $G$ for mixture generation.

In summary, our full objective for our framework is formulated as:
\begin{align}
&\min\limits_{E_n,G_c,G_m} \max\limits_{D_z,D_p}\mathcal{L}(E_n,G_c,G_m,D_z,D_p),
\label{eq:full}
\end{align}
where
\begin{align}
&\mathcal{L}(E_n,G_c,G_m,D_z,D_p)
\nonumber \\
=&\mathcal{L}_{content}(E_n,G_c,D_z)+\mathcal{L}_{mixture}(G_c,G_m,D_{p}).
\end{align}
In the above formulation, $\mathcal{L}_{content}$ corresponds to learning the content concept and $\mathcal{L}_{mixture}$ corresponds to the joint learning for mixture generator. 

{We refer to our framework as MIXture GAN (MIXGAN).} In the following, we elaborate each of them.

\subsection{Learning Content} 

We consider content as containing general spacial relationship of pixels, but without suffcient details in an image. Hence, we assume that content can be encoded by low-dimensional and abstract latent variables, and thus we model learning the content by an autoencoder structure with a bottleneck. On the other hand, as our objective is to learn the content concept for generation purpose, we would further like to force the latent variables to follow a prior, so that we can easily sample from it. Such a consideration leads us to an adversarial autoencoder (AAE) structure \cite{aae}, as shown in the upper left part in Figure \ref{fig:method}. A discriminator $D_z$ aims to distinguish the noise $z$ (sampled from a gaussian distribution $\mathcal{N}$) from the latent variables, and the encoder $E_n$ has an extra task: to foolish the discriminator by forcing the latent variables follow the same gaussian distribution $\mathcal{N}$.

Formally, let $\mathcal{L}_{content}$ be the loss function for learning content concept, and let $G_c$ be the content decoder in the AAE structure. We have our objective in learning content concept:
\begin{align}
&\min\limits_{E_{n},G_{c}} \max\limits_{D_{z}}\mathcal{L}_{content}(E_{n},G_{c},D_{z})\nonumber \\
=&\mathcal{L}_{adversarial}(E_{n},D_{z})+\lambda\mathcal{L}_{reconstruction}(E_{n},G_{c})\nonumber \\
=& \mathbb{E}_{z\sim \mathcal{N}}[(D_{z}(z)-1)^{2}]+\mathbb{E}_{x\sim p_{data}(x)}[(D_{z}(E_{n}(x)))^{2}] \nonumber \\
+&\lambda\mathbb{E}_{x\sim p_{data}(x)}[\parallel x-G_{c}(E_{n}(x))\parallel_{1}],
\label{eq:content}
\end{align}
where $x$ is an image from the content domain and $\lambda$ controls the relative importance of the reconstruction loss over the adversarial loss.
The $L_1$ reconstruction loss encourages the encoder to capture the low frequencies accurately but less high-frequency details than original images \cite{cyclegan}. For acquiring more stable learning, we use a least square form of the adversarial loss \cite{c30} instead of the negative log likelihood in the original GAN \cite{gan}.

\subsection{Mixture Generator} 

Now that we have learned some content concept in the AAE structure, we can propose our mixture generator $G$. We show the architecture in the right of Figure \ref{fig:method}. The mixture generator contains a content decoder $G_c$ and a mixture decoder $G_m$. The content decoder $G_c$ is also a part of AAE and thus has the ability to reproduce content concept by simply receiving noise sampled from $\mathcal{N}$. The content decoder $G_c$ provides the mixture decoder $G_m$ with rich hierarchical features which carry content concept of different levels.
The mixture decoder $G_m$ progressively takes these features as inputs (so it obtains content concept) and learns the style concept from outside (detailed in the next paragraph). The style concept is kept in the mixture decoder $G_m$'s architecture, and therefore the style concept can be ``released'' and join the content concept while $G_m$ is processing the hierarchical features. As a result, the mixture generator $G$ jointly learns both kinds of concepts and can generate images from a new domain.

Regarding learning the style concept, we consider style as information related to high-frequency details and local structures in local areas, e.g., color and sharpness. Its structural information can thus be independent across different patches of the whole image. Hence, we model learning the style by independently observing and learning from small patches of images. To this end, we employ a patch discriminator $D_p$ which looks at small patches of some candidate images and aims to distinguish the ones of the style domain from those generated by our mixture generator \cite{pix2pix}. The patch discriminator $D_p$ is illustrated in the lower left in Figure \ref{fig:method}. Here, $D_p$ is like a teacher passing style concept to a student, i.e., the mixture generator $G$, who absorbs the style concept in the mixture decoder $G_m$.

Formally, our objective of the mixture generator is:
\begin{align}
&\min\limits_{G_c,G_m} \max\limits_{D_{p}}\mathcal{L}_{mixture}(G,D_{p})=\mathcal{L}_{mixture}(G_c,G_m,D_{p})
\nonumber \\
=& \mathbb{E}_{y\sim p_{data}(y)}[((D_{p}(y)-1)^{2}]
+\mathbb{E}_{z\sim \mathcal{N}}[(D_{p}(G(z)))^{2}].
\label{eq:mixture}
\end{align}
where $y$ is an image from the style domain and $z$ is the noise sampled from the prior, i.e., gaussian distribution $\mathcal{N}$.

\subsection{Network Structure of the Mixture Generator}

As shown in Figure \ref{fig:method}, the content decoder consists of a block of three fully connected layers which process the high-level features, and a block of three convolutional layers which process mid-level and low-level features. Similar to \cite{pix2pix}, each of these layers is followed by a Batchnorm layer and a ReLU activation. The structure of mixture decoder $G_m$ is almost identical to the convolutional block of the content decoder $G_c$, as $G_m$ only captures mid-level and low-level local style concept, only except that it has one more convolutional layer to fully learn from the low-level features.
% \vspace{0.1cm}
\paragraph{Implementation details.} 
As the joint learning of the mixture decoder $G_m$ is based on the hierarchical features provided by the content decoder $G_c$. This is a two-step training process. We need to first train $G_c$ with images of the content domain. Therefore, we first optimize Eq. (\ref{eq:content}) in an adversarial learning pattern \cite{aae,gan}. During our training, we use Adam \cite{adam} solver with learning rate $0.0002$ and $\beta_1 = 0.5, \beta_2 = 0.999$.  It reaches convergence typically within 100 epoches.
Then, as $G_c$ can already ``reproduce'' the learned concept, we optimize Eq. (\ref{eq:mixture}) to train the mixture generator $G$, also in an iterative adversarial learning pattern. We use the same Adam solver and training typically converges within 300 epoches.
During inference stage, as the latent variables space (ideally) follows the prior, we can simply sample from the prior $\mathcal{N}$ for mixture generator.

% show content example
\begin{figure}[t]
\begin{center}
\includegraphics[width=0.8\linewidth]{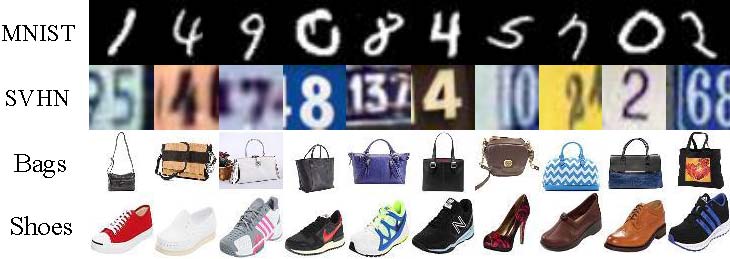}
\end{center}
   \caption{Samples in four training datasets.}
\label{fig:example}
\end{figure}
% first digit task's image
\begin{figure}[t]
\begin{center}
\includegraphics[width=0.8\linewidth]{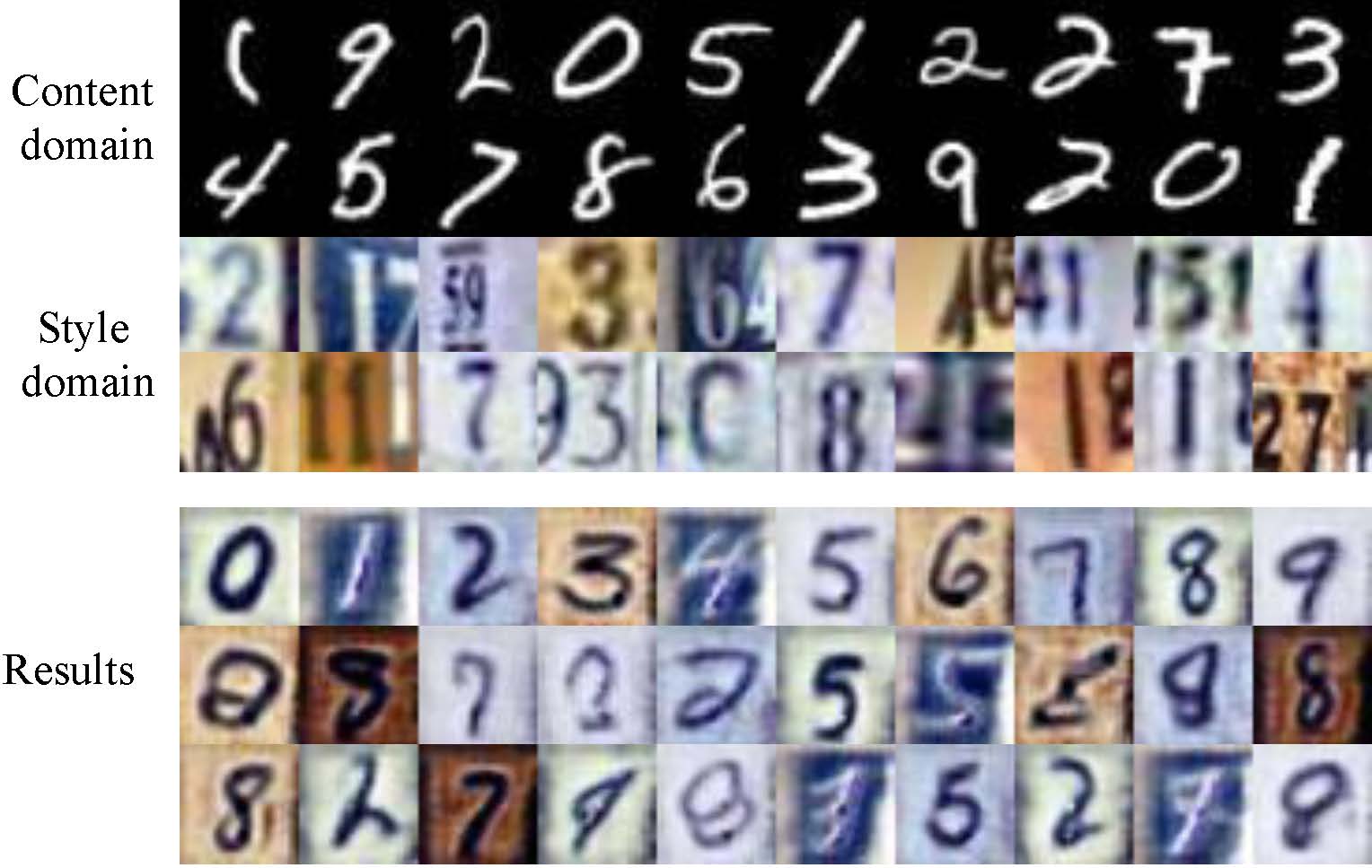}
\end{center}
   \caption{The first four rows: samples in MNIST (content domain) and SVHN (style domain). The last three rows: our results in mixture generation.}
\label{fig:digit1}
\end{figure}
% second digit
\begin{figure}[t]
\begin{center}
\includegraphics[width=0.8\linewidth]{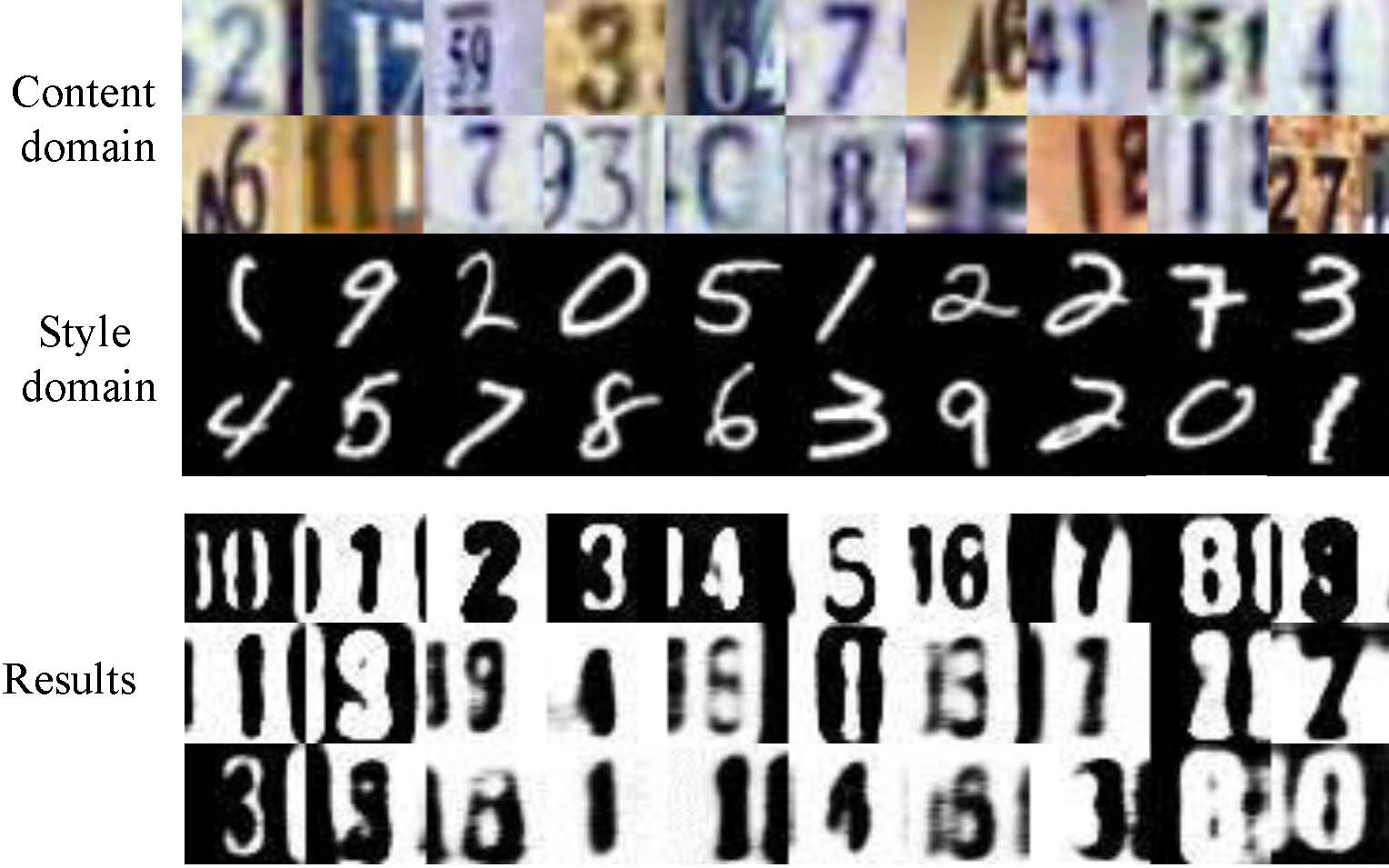}
\end{center}
   \caption{ The first four rows: samples in SVHN (content domain) and MNIST (style domain). The last three rows: our results in mixture generation.}
\label{fig:digit2}
\end{figure}
\section{Experiments}
We show experimental results on three aspects. Firstly we show that MIXGAN can achieve our goal, i.e., mixture generation. Then we show that in mixture generator, the content decoder and the mixture decoder learn content concept and style concept, respectively. Finally, we present comparative results with other related generative models to show that our framework performs better in mixture generation.

\subsection{Settings}

For evaluation of our framework as well as for comparison, we design two groups of experiments, with different content domains and style domains.

The first group focuses on digits. We take hand-written digits from the MNIST dataset \cite{mnist} and type-script digits from the SVHN dataset \cite{svhn}. For hand-written digits, the content refers to the relatively wild shapes, in contrast to the relatively uniform shapes of type-script digits. The style refers to their color schemes, where hand-written digits are black-and-white (BW) and type-script digits are colorful. We show some examples in the first two rows in Figure \ref{fig:example}.

%-----
The other group focuses on a more difficult task. We aim to learn content and style from bags and shoes, respectively, and in reverse. These images are available from \cite{pix2pix}. The content refers to the distinct shapes of both of them, and the style refers to the color style, as shown in the last two rows in Figure \ref{fig:example}.

Since we are targeting the mixture generation which typically involves generating a new domain, there is no groundtruth available for quantitative evaluation.
Thus, except visual results, we also design various proper methods including  human evaluations and quantitative evaluations for evaluating our framework as well as comparing with other methods.

% image for first task: grayscale bags + colorful shoes = colorful bags
%,height=0.18\linewidth
 \begin{figure}[t]
\begin{center}
\includegraphics[width=0.8\linewidth]{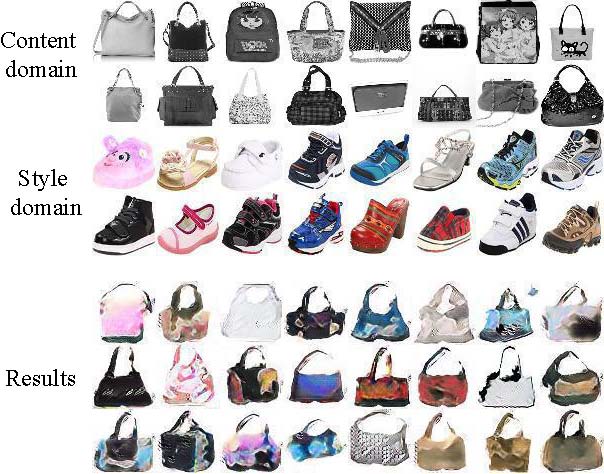}
\end{center}
   \caption{The first four rows: samples in grayscale bags (content domain) and shoes (style domain). The last three rows: our results in mixture generation.}
\label{fig:bag}
\end{figure}
% image for second task: grayscale shoes + colorful bags = colorful shoes
%,height=0.18\linewidth
\begin{figure}[t]
\begin{center}
\includegraphics[width=0.8\linewidth]{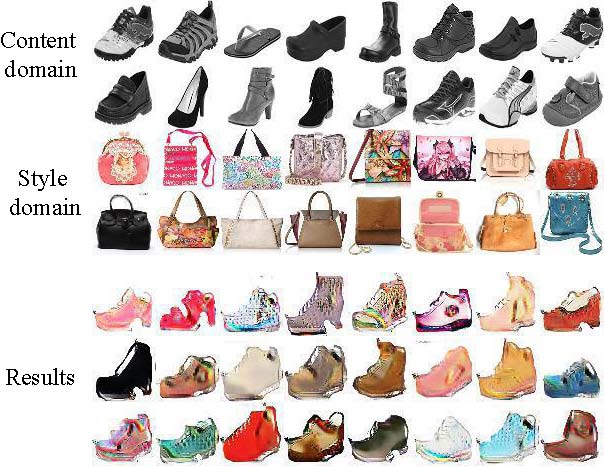}
\end{center}
   \caption{The first four rows: samples in grayscale shoes (content domain) and bags (style domain). The last three rows: our results in mixture generation.}
\label{fig:shoe}
\end{figure}

\subsection{MIXGAN for Mixture Generation}
In this section we show the results of MIXGAN for mixture generation.
  Successful mixture generation results account for above $70 \%$ in every task, so mixture generation is a big probability event for MIXGAN (please refer to Sec. 4.4 in details). Thus, we select successful samples to show the effectiveness of MIXGAN. 

% \vspace{0.1cm}

 \paragraph{Experiments on digits.}
Our first task of this experiment group is to learn to generate images with content of MNIST, i.e., handwritten digits, and style of SVHN, i.e., colorful style like SVHN. We show the results in Figure \ref{fig:digit1}. We can see from Figure \ref{fig:digit1} that MIXGAN can generate images containing hand-written digits while with a color style of the type-script ones. The generated domain is new as MIXGAN does not observe any images that look like the generated ones. Similar observations can be found in the mirror task, where the content is type-script digits and with BW color style. We show the results in Figure \ref{fig:digit2}.

%--------
% \vspace{0.1cm}
 \paragraph{Experiments on bags and shoes.}
Bags and shoes are a more difficult domain than digits since the objects are different and they are of higher resolution (64 by 64) compared to digits (32 by 32).
We show in Figure \ref{fig:bag} the experimental results where the content is from the bags and the style from shoes. Note that for better learning the content concept, we first transform the images of bags to grayscale images. This can be easily done and does not require considerable efforts. We can observe from Figure \ref{fig:bag} that, although the content and style are from different objects with higher resolution, MIXGAN can learn to absorb different kinds of concepts from them, and generate a new domain.

We also show the results in the mirror task in Figure \ref{fig:shoe}. We can see that MIXGAN can also learn content concept from the shoes and style concept from the bags.
%------------------------------------------------------

\subsection{Component Evaluation for Mixture Generator}
In this subsection, we evaluate how each component contributes to the MIXGAN. Specifically, we evaluate the components of the proposed mixture generator. Recall that in the mixture generator (Figure \ref{fig:method}), the content decoder serves to learn the content concept, and the mixture decoder learns the style concept as well as joins them for mixture generation. Here we take two tasks from the two experiment groups (one for each group) for example to illustrate the effects.

\begin{figure}[t]
\begin{center}
\includegraphics[width=1\linewidth]{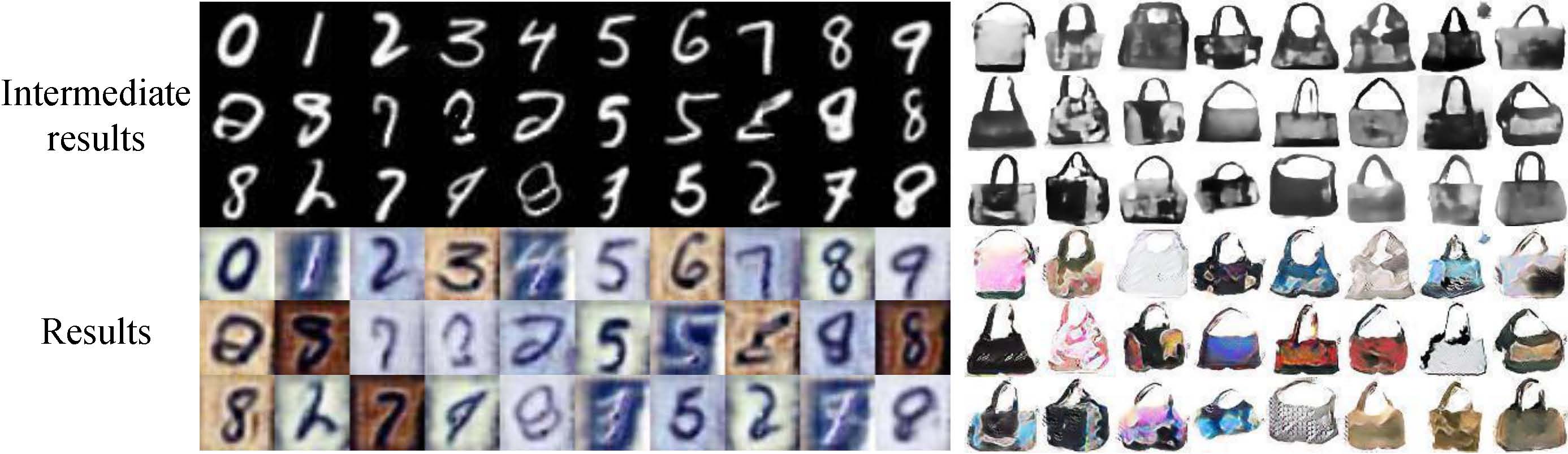}
\end{center}
   \caption{Left: results with MNIST (content domain) and SVHN (style domain) as training datasets. Right: results with grayscale bags (content domain) and shoes (style domain) as training datasets. The intermediate results: the content decoder outputs in the first three rows.  The final results: the mixture generator outputs in the last three rows. Corresponding pairs of samples share the same content in the lower and upper parts.
   }
\label{fig:content}
\end{figure}

We visualize the intermediate results produced by the content decoder, i.e., $G_c(z)$ where $G_c$ is the content decoder and $z$ is noise. We show these results in the upper part of Figure \ref{fig:content}. We can see that the intermediate digits already have the hand-written shapes, although without the desired color style. In contrast, the mixture generator produces images with desired color style, as shown in the lower part of Figure \ref{fig:content}. 
Similar observation can be found on the bags. This observation verifies that the content decoder can learn the content concept, while the mixture decoder can learn the style concept and join them for mixture generation.
% content image 
%----------------------Comparative experiment----------
% table for success rate
\begin{table}[h]
\begin{center}
\tiny
\setlength{\tabcolsep}{3.375pt}
\begin{tabular}{|c|c|c|c|c|}
\hline
Models & AAE &  LSGAN & CycleGAN & MIXGAN \\
\hline
TASK1 & $ 3.47 \% \pm 3.90\%$ & $2.82 \% \pm 1.56 \%$ & $ 2.54 \% \pm 2.63 \%$ & $ 85.63\% \pm 6.37\%$ \\
\hline
TASK2 & $ 1.17 \% \pm 1.64\%$ & $7.32 \% \pm 5.33 \%$ & $ 0.39 \% \pm 0.72 \%$ & $ 71.13\% \pm 7.53\%$ \\
\hline
TASK3 & $ 1.95 \% \pm 2.33\%$ & $0.25 \% \pm 0.71 \%$ & $ 0.00 \% \pm 0.00 \%$ & $ 86.17 \% \pm6.62 \%$ \\
\hline
TASK4 & $1.95 \% \pm 1.62 \%$ & $0.50 \% \pm 0.92 \%$ & $ 0.00 \% \pm 0.00 \%$ & $ 90.70 \% \pm4.67 \%$ \\
\hline
\end{tabular}
\end{center}
\caption{Success rate evaluated by human annotators. Measured by average rate $\pm$ std. Task 1: generating colorful handwritten digits (training datasets: MNIST, SVHN). Task2: generating black-and-white  type-script digits (training datasets: MNIST, SVHN). Task3: generating colorful bags (training datasets: grayscale bags and colorful shoes). Task4: generating colorful shoes (training datasets: grayscale shoes and colorful bags).}
\label{table:rate}
\end{table}

% table for comparing our model with LSGAN, AAE, cycleGAN
\begin{table}[h]
\begin{center}
\tiny
\setlength{\tabcolsep}{3.375pt}
\begin{tabular}{|c|c|c|c|c|}
% \hline
\hline
Models & AAE &  LSGAN & CycleGAN & MIXGAN \\
\hline
MNIST & $22.47 \pm 1.06$ & $21.46 \pm 0.30$ & $48.98 \pm 2.10 $ & $ 248.11\pm 1.88$ \\
\hline
SVHN & $103.48 \pm 1.34$  & $102.17 \pm 4.84$ & $17.47 \pm 0.78$ & $251.96 \pm 3.44$ \\
\hline
\end{tabular}
\end{center}
\caption{MMD distances of the generated samples and the two training sets in Task 1: generating colorful handwritten digits (training datasets: MNIST, SVHN). The MMD distances are computed using the feature from the last hidden layer of the pretrained deep binary classifier network.}
\label{table:mmd}
\end{table}

\subsection{Comparison to Related Models}

% image for comparing models when training datasets are the MNIST dataset and the SVHN dataset
\begin{figure}[t]
\begin{center}
\includegraphics[width=0.8\linewidth]{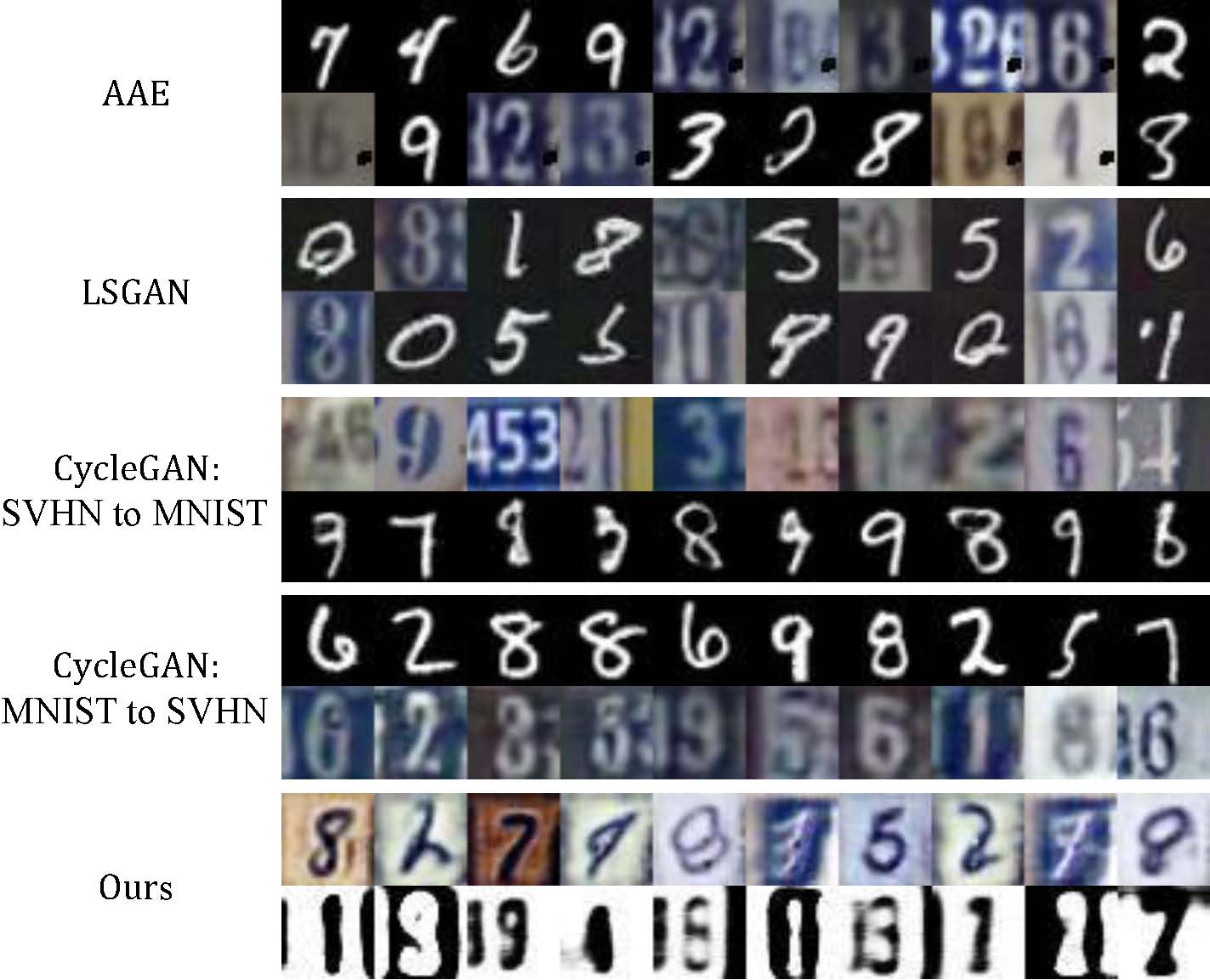}
\end{center}
   \caption{ Comparative experiment results. Training datasets: the MNIST dataset and the SVHN dataset.}
\label{fig:comdigit}
\end{figure}

% tne 
\begin{figure}[t]
\begin{center}
\includegraphics[width=0.8\linewidth]{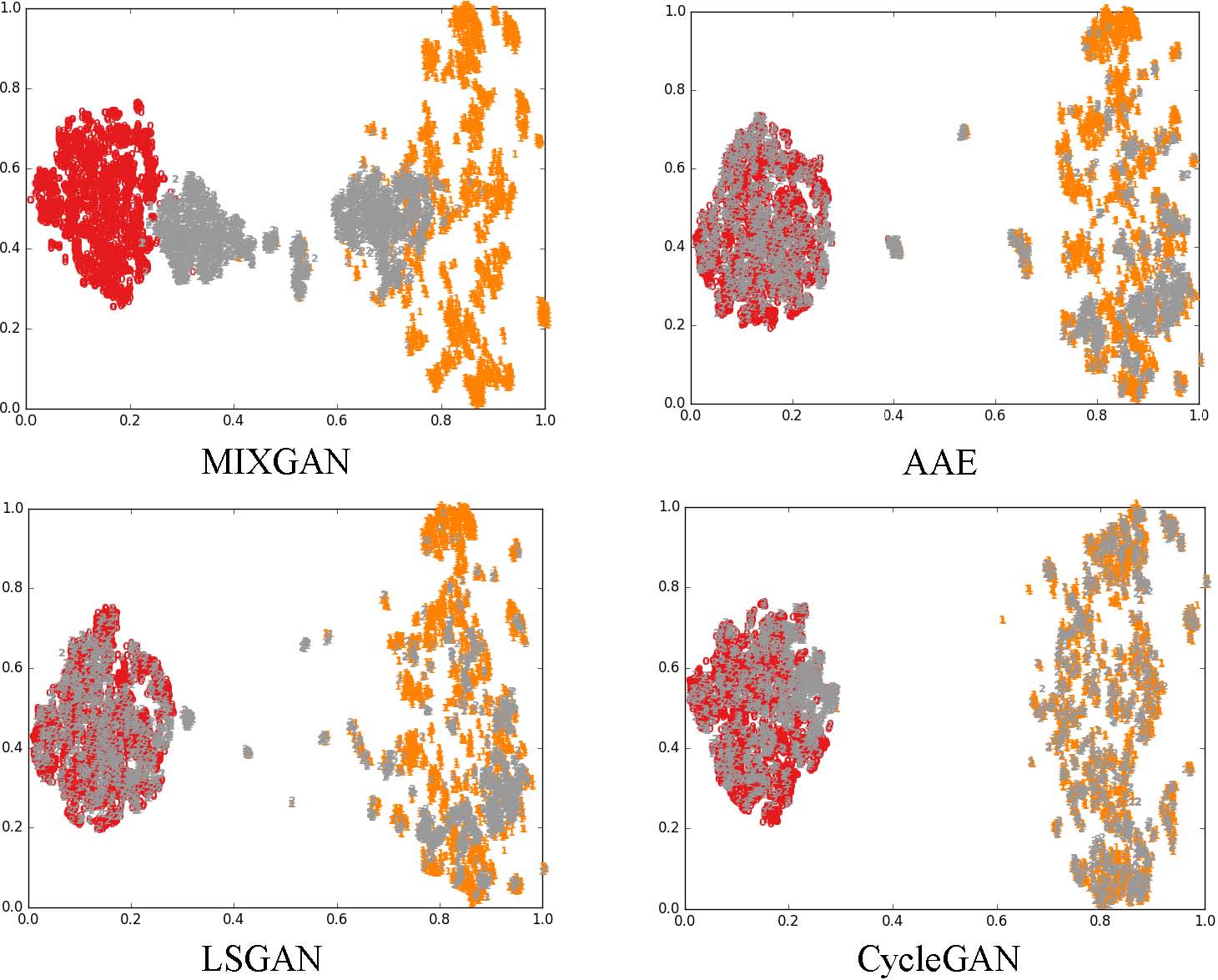}
\end{center}
   \caption{Visualization of the relationships between the generated samples and the training samples in generating colorful hand-written digits (Task 1 in Table 1). The red, yellow and grey points correspond to samples from MNIST, SVHN and the generated ones, respectively}
\label{fig:tsne}
\end{figure}
% image for annotating
\begin{figure}[t]
\begin{center}
\includegraphics[width=0.8\linewidth]{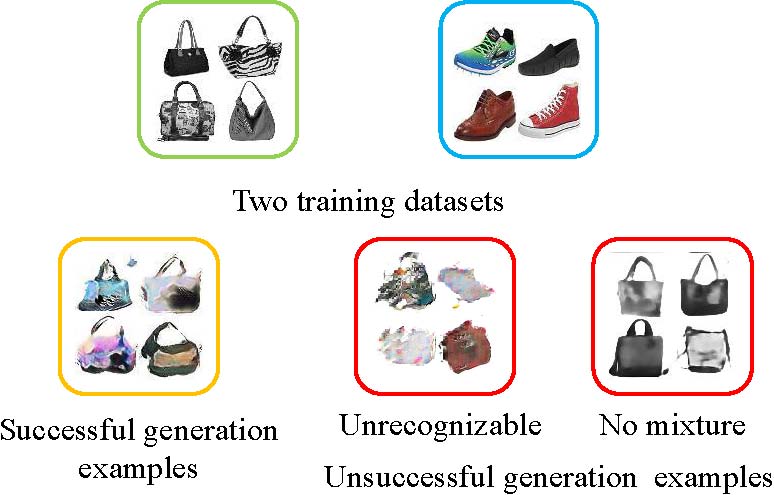}
\end{center}
   \caption{Illustration of the criteria of judging successful mixture generation by human annotators.}
\label{fig:npexample}
\end{figure}
%----------

Now we compare MIXGAN with several related generative models by comprehensive visual and quantitative evaluation, including GAN-based generation models LSGAN \cite{c30}, adversarial autoencoder (AAE) \cite{aae} and a state-of-the-art image translation model cycleGAN \cite{cyclegan}.

%--------
% To adapt to the mixture generation problem, we use the same training sets as ours for these models. Since they are designed for conventional generation problem which learns from a single domain, we combine the two domains (datasets) for them. We show the comparative results on the digit tasks in Figure \ref{fig:comdigit}, and show the results on the bag-shoe task in Figure \ref{fig:npexample}.

% \noindent \textbf{Comparison with GAN-based models}. From Figure \ref{fig:comdigit} we can see that the conventional generation models, i.e., AAE and LSGAN, can learn to generate images within a specific domain, either BW hand-written digits or colorful type-script ones. However, they could not generate a new domain for mixture generation, \Kr{because these models are not aware of different domains and learn from them without distinction of different concepts.} 
% In contrast, MIXGAN \Kr{has specifically designed strategies to learn content concept, style concept and how to join them, respectively. Therefore, MIXGAN can learn different \ja{concepts} from the two domains for mixture generation (i.e., colorful type-script digits)}. 
%--------
% \vspace{0.1cm}

\paragraph{Visual results.}
From Figure \ref{fig:comdigit} we can see that the conventional generation models, i.e., AAE and LSGAN, can learn to generate images within a specific domain, either BW hand-written digits or colorful type-script ones. However, they could not generate a new domain for mixture generation. We can also observe that although cycleGAN can translate an image of hand-written digit to another of type-script digit, the content and style remain the same as in their original domains, while MIXGAN can achieve the mixture generation. In contrast, MIXGAN has specifically designed strategies to learn content concept, style concept and how to join them, respectively. Therefore, MIXGAN can learn different concepts from the two domains for mixture generation (i.e., colorful type-script digits).

To further explore the differences, we evaluate the mixture generation by visualizing the distributions of both the training samples and the generated samples using t-SNE \cite{tsne}. We show the comparative results in the task of generating colorful hand-written digits in Figure \ref{fig:tsne} (corresponding to the Task 1 results in Table \ref{table:rate}). We can see that MIXGAN is able to generate samples belonging to a distribution which is ``between'' the two training distributions. In contrast, the compared models fail in this mixture generation task. They only generate samples belonging to the original training distributions.

% \vspace{0.1cm}

 \paragraph{Human evaluations.}
We organize 100 human annotators to evaluate success rate of MIXGAN and the compared models in the mixture generation task. As our objective is mixture generation, we define that a generated image is ``successful'' if it can be recognized as in a new domain combining the content and style from the two training domains, respectively. Here the key criteria are recognizability and combination/mixture of content and style. For example, we show in Figure \ref{fig:npexample} some typical successful and failed cases in the task of combining the content of shoes and style of handbags. For each of our evaluation tasks (4 in total), each human annotator judges 200 samples generated by MIXGAN whether successful or not. The same evaluation is performed on the compared models, namely AAE, LSGAN and CycleGAN. We show the comparative results in Table \ref{table:rate}. The average success rate of MIXGAN is above $70\%$, while the second best is always below $10\%$.

% \vspace{0.1cm}

 \paragraph{Quantitative evaluations.}
Although we can intuitively recognize how well the models can achieve mixture generation in a 2-D embedding in Figure \ref{fig:tsne}, we further evaluate it quantitatively.
To this end, we adopt the maximum mean discrepancy (MMD) between the generated samples and the training samples. Smaller MMD distance suggests that the generated sample still belong to one of the original training distributions. But we note that, larger MMD distance between the generated samples and training samples does not directly suggest a good mixture generation performance (because it can just fail to generate meaningful samples). Therefore, this measure should be considered complementary to the human annotator success rate: higher success rate with a higher MMD distance suggests better mixture generation.

To compute the MMD distance, we need to first of all determine which training distribution a given sample is compared with, because there are two distinct training distributions. To achieve this, we first assign each generated sample to a ``nearer'' training set, and then compute the MMD distance between the samples and their corresponding training set. We define ``near'' by training a deep binary classifier network which is trained using the two training sets and produces a probability that a sample is belonging to a specific training set. Then we use this classifier to judge to which training set each generated sample should be assign.

We show the comparative results in Table \ref{table:mmd}. Together with Table \ref{table:rate}, we can see that MIXGAN has higher success rate and MMD distance, indicating the better performance in terms of mixture generation.

% \noindent \textbf{Comparison with image translation model}. From Figure \ref{fig:comdigit}, we can observe that although cycleGAN can translate an image of hand-written digit to another of type-script digit, the content and style remain the same as in their original domains, while MIXGAN can achieve the mixture generation. The difference comes out because the translation models can only translate images in one domain to another known domain, whereas our framework can learn concepts from different domains for mixture generation in a new domain beyond the observed ones.

In summary, as compared to AAE, LSGAN and cycleGAN,
our MIXGAN model can learn to generate samples belonging to a new domain with concepts absorbed from different domains.

\section{Conclusion}
We present our work on mixture generation, where we aim to generate a new domain beyond the training ones. In particular, we develop MIXGAN to generate a new domain that contains the content and style concepts extracted from two different domains. The core of MIXGAN is the mixture generator, which jointly learns two kinds of concepts as well as how to join them for mixture generation. Our experiments have shown its successful applications as compared to the sate-of-the-art GAN-based methods.

\section*{Acknowledgements}
This work was supported partially by the National Key Research and Development Program of China (2018YFB1004903), NSFC (61522115, 61661130157, 61472456, U1611461), Guangdong Province Science and Technology Innovation Leading Talents (2016TX03X157), and the Royal Society Newton Advanced Fellowship (NA150459).

Finally we thank the anonymous reviewers for their invaluable comments.

\bibliographystyle{named}
\bibliography{ijcai18}

\end{document}